\documentclass[letterpaper, 10 pt, conference]{ieeeconf}  

\IEEEoverridecommandlockouts                              

\usepackage{color}
 \usepackage{booktabs}
\usepackage{times}
\usepackage{epsfig}
\usepackage{graphicx}
\usepackage{tabularx}
\usepackage{amsmath}
\usepackage{amssymb}
\usepackage{bbm}
\usepackage[misc]{ifsym}
\usepackage[export]{adjustbox}
\usepackage{rotating}

\usepackage[ruled,vlined,linesnumbered]{algorithm2e}
\usepackage{graphicx} 

\usepackage{subfigure}
\usepackage{tikz}
\usepackage{verbatim}
\usepackage{booktabs}
\usepackage{multirow}
\usepackage{makecell}
\usepackage{authblk}
\usepackage{hyphenat} 

\usepackage{pgf}
\usepackage{algpseudocode}
\usetikzlibrary{arrows,automata}

\usepackage{graphbox}

\makeatletter
\newcommand{\removelatexerror}{\let\@latex@error\@gobble}
\makeatother

\makeatletter 
  \newcommand\figcaption{\def\@captype{figure}\caption} 
  \newcommand\tabcaption{\def\@captype{table}\caption} 
\makeatother

\makeatletter
\let\NAT@parse\undefined
\makeatother
\usepackage[colorlinks]{hyperref}  

\usepackage{graphics}
\usepackage{pifont}
\newcommand{\cmark}{\ding{51}}%

\title{ Analysis of Exploration vs. Exploitation in Adaptive Information Sampling}
\author{Aiman Munir \and Ramviyas Parasuraman \thanks{The authors are with the Heterogeneous Robotics Research Lab, Department of Computer Science, University of Georgia, Athens, GA 30602, USA. 
Email: {\it \{am68149,ramviyas\}@uga.edu}. }}

\begin{document}

\newtheorem{definition}{Definition}
\newtheorem{theorem}{Theorem}
\newtheorem{lemma}{Lemma}
\newtheorem{proposition}{Proposition}
\newtheorem{property}{Property}
\newtheorem{observation}{Observation}
\newtheorem{corollary}{Corollary}

\maketitle
\thispagestyle{empty}
\pagestyle{empty}

\begin{abstract}
Adaptive information sampling approaches enable efficient selection of mobile robot's waypoints through which accurate sensing and mapping of a physical process, such as the radiation or field intensity, can be obtained. 
This paper analyzes the role of exploration and exploitation in such information-theoretic spatial sampling of the environmental processes. We use Gaussian processes to predict and estimate predictions with confidence bounds, thereby determining each point’s informativeness in terms of exploration and exploitation. Specifically, we use a Gaussian process regression model to sample the Wi-Fi signal strength of the environment. For different
variants of the informative function, we extensively analyze and evaluate the effectiveness and efficiency of information mapping through two different initial trajectories in both single robot and multi-robot settings. The results provide meaningful insights in choosing appropriate information function based on sampling objectives.

\end{abstract}

\section{Introduction}

Mobile robot-aided mapping of environmental processes, such as information sampling \cite{OnlineInfoGathering}, sensor coverage \cite{AdaptiveSamplingSensorCoverage}, localization of source \cite{Fink2010}, and monitoring of environmental phenomena \cite{LiuK2017}, have been well investigated. Particularly, sensor coverage with multiple robots involves positioning robots in an optimal manner to maximize overall performance in terms of sensing the environmental phenomenon. Monitoring is a persistent process for identifying anomalies in physical processes by efficiently collecting the most informative samples. For all of these objectives, modeling of the underlying processes is required or involved, which can be obtained via sampling or mapping. 

Modeling of physical processes plays an important role in autonomous robots’ decision-making. Robots need to create a model of the environmental phenomenon to accomplish mapping tasks, especially when the environment is unexplored. Mapping spatial distribution enables the robots to work autonomously in search and rescue missions and make decisions without human intervention (e.g., for rescuing targets in areas with high radiation exposure). Similarly, the robot requires knowing the areas with higher risks so that the robot can always choose a path to connect to the network and maintain communication, etc. \cite{caccamo2017rcamp,parasuraman2013spatial}.
Therefore, robotics researchers are actively investigating different strategies for mapping physical processes, such as radiation, Wi-Fi signal strength, gas distribution, radio signal strength etc., using unmanned vehicles \cite{AdaptiveSampling, Fink2010,Stachniss2009}. 

In order to map the environmental phenomenon, it is crucial to measure the value of a physical process throughout the environment. However, not all the locations can provide useful information about the change in the process itself.  Primary considerations involved in mapping such processes include the degree of autonomy, accuracy, and efficiency. Measuring intensity at every location is deemed impractical; hence, dense sampling is not viable in mapping. Instead, an accurate, time- and cost-effective process model can be obtained by gathering samples from the points that contain the most significant amount of information.

Exploration refers to accumulating samples from previously unexplored areas to reduce uncertainly in the map, while exploitation implies determining the next sampling point based on the best information from the current estimates (to localize source, for example). 
Mapping algorithms and techniques in the literature either use exploration or exploitation, or alternate between exploration and exploitation as needed. In \cite{Fink2010}, an active control law for mobile robot is proposed to shift between exploration and exploitation objective. In \cite{AdaptiveSamplingSensorCoverage}, a utility function is used for adjusting between exploration and exploitation. On the other hand, parallel strip routes are used in \cite{Zakaria2017, Gabrlik2020} to explore the environment for mapping of the spatial distribution. These two techniques are fundamental to the mapping process, so the purpose of the study in this paper is to compare how well various exploration and exploitation techniques perform, as well as analyze how trade-offs between exploring and exploitation affect mapping. 
Specifically, the contributions made in this paper are three-fold.
\begin{itemize}
    \item We comprehensively compare the adaptive information sampling techniques in terms of exploration and exploitation objectives from a perspective of mapping physical processes using Gaussian Processes \cite{williams1996gaussian,Fink2010}.
    \item We systematically analyze the impact of the information function on the mapping accuracy and efficiency through extensive simulations in single robot experiments with random walk and fixed sweep trajectories as the baselines for comparison.
    \item We extend this analysis to multi-robot settings with fixed and dynamic Voronoi partition-based adaptive sampling \cite{kemna2017multi} assignments to each robot in the system.
\end{itemize}
The results from this study provide meaningful insights to choose appropriate sampling techniques to balance between the exploration (minimizing uncertainty) and exploitation (maximizing information value) from the map estimates through adaptive sampling with mobile robots.

\section{Related Work}

\begin{table*}[ht]
\caption{\label{tab:Literature Review} 
Comparison of related works in the literature on adaptive information sampling. The Information functions used to obtain waypoints for successive sampling are adapted from the respective references in this table.}
\resizebox{\textwidth}{!}{%
\begin{tabular}{|p{1cm}|p{5.5cm}|p{1cm}|p{1cm}|p{2cm}|p{2.2cm}|p{2.2cm}|p{2.2cm}|p{2.2cm}|p{2.2cm}|}
\hline

Reference & \textbf{Information Function} & \multicolumn{2}{c|}{\textbf{Scalability}}& \textbf{Robot type} & \textbf{Sampling Type} & \textbf{Objective} & \textbf{Property/Process Measured}  & \textbf{Prediction Model} & \textbf{Exploration or Exploitation}  \\ \hline
            
& \textbf{}  & \textbf{Single Robot} & \textbf{Multi Robot} & \textbf{}   & \textbf{}  & \textbf{}  & \textbf{}   & \textbf{} &\textbf{}                                                       \\ \hline

\textbf{\cite{OnlineInfoGathering}}   & $I\left(\mathcal{P}_{\mathbf{x}, \mathbf{x}^{\prime}}\right) = \frac{1}{\left|\mathcal{P}_{\mathbf{x}, \mathbf{x}^{\prime} \mid}\right|} \sum_{\tau \in \mathcal{P}_{\mathbf{x}, \mathbf{x}^{\prime}}} H({\tau})$ & \cmark                   &                    & Ground        & Discrete       & Mapping                                   & Magnetic Field Intensity    & Gaussian Processes       & Exploration                                          \\ \hline

\textbf{\cite{Fink2010}} &  $\mathcal{I}_{i, j}=-\frac{\partial^{2} \mathcal{L}\left(\theta_{m}\right)}{\partial \theta_{m, i, j}}$ & \cmark     & \cmark   & Ground     & Discrete & Mapping and source localization   & Radio Signal Strength    & Gaussian Processes    & Mix  \\ \hline

\textbf{\cite{LiuK2017}}                           & $I\left(Z_{A} ; Z_{B}\right)=I\left(Z_{B} ; Z_{A}\right)=H\left(Z_{A}\right)-H\left(Z_{A} \mid Z_{B}\right)$                                    & \cmark                   &                    & Marine              & None                                         & Monitoring                    & Salinity                                                          & Sparse Gaussian Processes       &   Exploration                                       \\ \hline

\textbf{\cite{caccamo2017rcamp}}                           &    $I^{rssi}(x) = \frac{max(\mu) - \mu*(x)}{max(\mu) - min(\mu) + \epsilon} $                               & \cmark                   &                    & UGV              & Continuous                                         & Path Planning                    & Wi-Fi                                                          & Gaussian Processes       &   Exploitation                                       \\ \hline

\textbf{\cite{Zakaria2017}}   & None   & \cmark                   &              & Ground  & Discrete (zigzag waypoints sweeping pattern) & Mapping   & Gamma radiation   & None    & Exploration                                                         \\ \hline

\textbf{\cite{AdaptiveSampling}} & $I({x})=\ln (\sigma \sqrt{2 \pi e}) $ 
& \cmark                   & \cmark                  & Ground               & Discrete             & Mapping          & Radio Signal Strength            & Gaussian Processes   & Exploration                                              \\ \hline

\textbf{\cite{Gabrlik2020}}                & None                                                &                     & \cmark                  & Ground and Aerial   & Continuous                                   & Localization of Sources                   & Radiation                                                          & None & Exploration                                                             \\ \hline

\textbf{\cite{kemna2017multi}}   & $\mathbb{H}\left[\mathbf{Y}_{x_{i+1}} \mid d_{i}\right]=\log \sqrt{2 \pi e \sigma_{Z_{x_{i+1} \mid d_{i}}}^{2}}+\mu_{Z_{x_{i+1} \mid d_{i}}}$        &            &  \cmark        &Underwater  & Discrete    &Modeling   &Algae    &Gaussian Processes     &Mix  \\ \hline

\textbf{\cite{Luo2019}}      & $ I(x)=\mu_{x \mid \tilde{V}_{i}, \mathrm{y}_{i}}^{*}+\beta \sigma^{* 2} _{x \mid \tilde{V}_{i}, \mathrm{y}_{i}}  $
&                     & \cmark                  & Ground & Discrete          & Sensing Coverage      & Stalk count , Temperature         & Mixture of Gaussian Processes & Mix                                   \\ \hline

\textbf{\cite{Juan2016}} & None    &                     & \cmark                  & Ground and Aerial   & Discrete       & Mapping/ Environmental Monitoring         & Temperature, humidity, luminosity and carbon dioxide concentration &    None   & Exploration                              \\ \hline

\textbf{\cite{Nur2020}}         & None            & \cmark                   &                    & Ground              & Discrete        & Mapping                                   & Radiation                                                          & None & Exploration                                                             \\ \hline

\textbf{\cite{Duvallet2008}}  & None         & \cmark         &                    & Ground              & Continuous                                   & Localization of Source/Vehicle            & Wi-Fi     & Gaussian Processes  & Exploration                                               \\ \hline

\textbf{\cite{Miyagusuku2016}}     & $p\left(z_{n e w j} \mid \mathbf{x}_{*}\right)=\Phi\left(\frac{z_{n e w, j}-{E}\left[z_{* j}\right]}{{var}\left(z_{* j}\right)}\right)$
  & \cmark                   &                    & Ground              & Discrete      & Robots localization                       & wireless signal strength                                           & Path loss and Gaussian Processes  & Exploration                           \\ \hline

\textbf{\cite{Cortez2008}}        & None / $\varphi=\gamma_{d}+1 / e^{{\beta_{0}}^{1/k}}$ & \cmark                   &                    & Ground              & Discrete/Continuous                          & Mapping                                   & Radiation                                                          & None & Exploration\\ \hline

\textbf{\cite{Newaz}}  & $ I:=a b s\left(H\left(z_{1: t}\right)-H\left(z_{1: t} \mid x_{t}\right)\right) $   & \cmark                   & None                & Discrete                                     & Localization of Hotspot    &                & Radiation & None
& Explore                                                            \\ \hline

\textbf{\cite{adaptive}}        & $ I(x)=\mu_{x \mid \bar{V}, \mathbf{Y}}^{*}+\beta \sigma_{x \mid \tilde{V}, \mathbf{Y}}^{*^{2}} $  &                     & \cmark                  & Ground and Aerial   & Discrete                        & Mapping                                   & Wi-Fi                                                               & Mixture of Gaussian Processes      & Mix                                \\ \hline

\textbf{\cite{cortez2008radiation}}                        & None                                        &                     & \cmark & Ground              & Continuous                                   & Mapping                                   & Gamma radiation                                                    & None       & Exploration                                                      \\ \hline

\textbf{\cite{Jilek2015}}               & None                                                &                     & \cmark                  & Ground and Aerial   & Discrete / Continuous                          & Localization                              & Gamma radiation                                                    & None        & Exploration                                              \\ \hline

\end{tabular}
}
\end{table*}

Both adaptive (taking informativeness of the sampled data into account) and non-adaptive (without considering informativeness) sampling approaches have been previously reported in the literature. Non-adaptive sampling methods focus on sampling the whole environment \cite{Gabrlik2020,Nur2020,Cortez2008}. 
Although, non-adaptive methods are less time-consuming, but in such methods it is hard to achieve the desired threshold of certainty. On the other hand, adaptive sampling methods provide convergence to an objective (threshold) and sampling can be stopped as soon as the desired threshold is achieved. The approach of planning a robot path that contains most-informative samples has been studied in the past. 

In \cite{OnlineInfoGathering}, a technique that maximizes the mean entropy information metric is used to search a station. Then using an RRT based Informative Planner algorithm it selects the path that provides maximum utility, i.e., trades off the informativeness and cost to reach that station. Authors in \cite{LiuK2017} uses entropy as an information criterion over Sparse Gaussian Process to find the most informative locations to persistently monitor salinity in the ocean. Similarly, \cite{Miyagusuku2016} uses wireless signals for robot localization in an indoor GPS-denied environment. It learns a path loss model from the data and then training the Gaussian process with the mismatches between the models and the data with a focus on better model variance prediction. In \cite{AdaptiveSampling}, the authors focused on mapping in structured environments. The algorithm partitions the environment for each robot and uses differential entropy as an information theory metric on top of Gaussian Process predictions to determine the next sampling point.
The authors in \cite{Newaz} proposed a Hexagonal Tree (HexTree) based sampling algorithm, which takes samples over a set of hexagonal  grid  points  and  builds  a  tree  of  possible trajectories  by  extending  candidate  trajectories  toward  the sampled  points.

Table~\ref{tab:Literature Review} provides detailed information about closely related works in the literature on adaptive information gathering. According to the table and literature review, several objective functions have been used in coordination with Gaussian Process Regression \cite{williams1996gaussian}  to map the physical process (i.e., to predict the samples at unvisited (unexplored) locations with confidence bounds). Despite a plethora of research employing different exploration and exploitation approaches, either on their own or in conjunction with their information function, to the best of our knowledge, no analysis exists in the literature on how much impact these techniques have on the effectiveness of mapping the underlying spatial distribution. Therefore, we assess how exploration and exploitation in combination with alternatives of information function influence the mapping process.

\section{Gaussian Processes Aided Robot Sampling}\label{AA}
Gaussian processes (GP) have been utilized for modeling spatial process. For example, in \cite{Gu2012}, Gaussian process regression (GPR) was used to model spatial functions for mobile wireless sensor networks and to generate a likelihood model for signal strength measurements. An algorithm in \cite{Duvallet2008} is presented to localize Wi-Fi using Gaussian Process and then estimates the global position of an autonomous vehicle in an industrial environment. The authors in \cite{Miyagusuku2016} used path loss to learn a model from data and then feed the mismatches to GP. In \cite{Fink2010}, GPR is used to obtain a model of radio signal strength, which in turn is used to get maximum likelihood of the source location. For the purpose of obtaining a spatial map in an environment with limited communication, \cite{kemna2017multi} used GPR to model algae bloom.

GP is a non-parametric continuous functions. It defines a probability distribution over functions. It assumes that every point has a normal distribution and there is a correlation between values at these points. It not only predicts a value, but also provides confidence bounds for the given measurements. 
Let $q$ be the  location from where the signal strength is measured and $z$ be the measurement, and let $q'$ be the testing points. The value of $z$ at any point $i$ is
\begin{equation}
z_{i}=f\left({q}_{i}\right)+\varepsilon ,
\end{equation}
where $\varepsilon$ is an additive Gaussian noise. We are interested in calculating a function $f(q')$ that makes predictions for given test locations $q'$. GP generates the following mean and variance for the test points $q'$. With every sample $q$ added to the training set, the values of $q'$ are updated accordingly. 
\begin{equation}
\mu_{z'}\left( {q}, {q'}, z\right)=k\left({q'}, {q}\right) k(q, q) z 
\label{eqn:mean}
\end{equation}
\begin{equation}
\sigma_{z '}^{2}\left({q},{q'}\right)=k\left({q'}, {q'}\right)-k\left({q'}, {q}\right) k({q}, {q})^{-1} k\left({q}, {q'}\right) 
\label{eqn:var}
\end{equation}
Here, $k(q,q')$ is the covariance function(kernel) and defines the correlation between $q$ and $q'$. One of the widely used kernel is squared exponential.
\begin{equation}
k\left( {q},  {q'}\right)=\sigma_{f}^{2} \exp \left(-\frac{1}{2 l^{2}}\left| {q}- {q'}\right|^{2}\right)
\end{equation}
where $q$ and $q'$ are two training samples, $q,q'\epsilon  R^2$, $\sigma_f$ are l are hyperparameters of GP and are called variance and length respectively. These hyperparameters are learned from the training data. The variance and mean in Eq.~\eqref{eqn:mean} and \eqref{eqn:var} are used to calculate the informativeness of every point.

\section{Adaptive Information Sampling}
\label{sec:information}
A method for determining the next sampling point using a utility function during the model creation process is termed adaptive information sampling \cite{Singh2009, AdaptiveSampling}. 
Depending on the criteria for selecting the next sampling location, adaptive sampling can be exploration-based, exploitation-based, or a mixture of both types.
In this paper, we have used two non-adaptive sampling baseline approaches that do not use an information function: a predefined sweep trajectory on a specific pattern; and a random walk. For adaptive information sampling, the following Information (utility) function to calculate the informativeness at every point $q$: 
\begin{equation}
I(q) = \alpha\mu_q+\beta\sigma_q^2 ,
\label{eqn:information}
\end{equation}
where $I(q)$, \(\mu_q\) and \(\sigma_q^2\) denote the informativeness, mean and variance of the point $q$. While \(\alpha\) and \(\beta\) are the importance factor for mean and variance, respectively, and determine the weightage given to mean and variance to calculate informativeness of point q.
Based on different weighted combinations of mean and variance in the informative function, we used the following approaches with baselines:
\begin{enumerate}
\item MaxMean - MaxMean approach chooses the point with maximum intensity value, i.e., max mean location as the next sampling point.
\item Alpha0.75 - Alpha0.75 approach selects the location with the highest information value given $\alpha = 0.75$.
\item Alpha0.5 - Alpha0.5 method selects the location with the highest information value given $\alpha = 0.5$.
\item Alpha0.25 - Alpha0.25 approach selects the location with the highest information value given $\alpha = 0.25$.
\item MaxVar - MaxVar chooses the location with the lowest confidence value as the target location.
\item MaxVarMaxMean - MaxVarMaxMean first selects the points with maximum uncertainty until a given threshold for confidence (variance) is reached. After satisfying the threshold, it then selects the points with MaxMean as the point of interest.
\end{enumerate}

\begin{table}[]
\caption{Values of alpha and beta for different Information $I(q)$ variants.}
\label{tab:alphabeta}
\centering
\begin{tabular}{|c|c|c|}
\hline
Approach  & Alpha & Beta \\ \hline
MaxMean   & 1     & 0    \\ \hline
Alpha0.75  & 0.75  & 0.25 \\ \hline
Alpha0.5  & 0.5   & 0.5  \\ \hline
Alpha0.25 & 0.25  & 0.75 \\ \hline
MaxVar    & 0     & 1    \\ \hline
\end{tabular}
\end{table}

Table~\ref{tab:alphabeta} enlists the values of alpha and beta for different variants used in this study. We have used these values to evaluate the effects of different levels of exploration and exploitation. As per Table~\ref{tab:Literature Review}, most utility functions in informative sampling rely on the mean, variance, or some combination of the two. Because of diversity, we have not used the same weights as in literature, but instead used a range of weights that can give a general idea of how a fix increase in weights can affect mapping performance. Nevertheless, this analysis can also be used to shed light on a specific scenario of exploration and exploitation.

\section{Experiment Design and Implementation}

\begin{figure}[ht]
    \centering  
  \includegraphics[width=0.98\columnwidth]{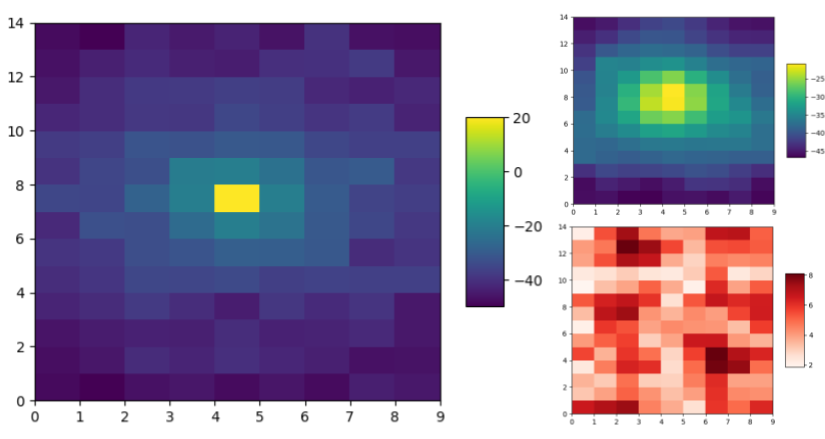}
   \caption{An image of the Wi-Fi signal ground truth (left) at source location (4,7) and predicted GP mean (top right) and variance (bottom right) of a simple random walk exploration strategy.}
   \label{fig:wifi}
\end{figure}

We have developed the simulations using the Robot Operating Systems (ROS \cite{quigley2009ros}) Gazebo simulation framework, built on top of the open-source code base from \cite{adaptive}. 
We considered a 10 m $\times$ 15 m simulated area free of obstacles (to avoid bias in the analysis due to collision avoidance algorithms). Until the robot's battery is depleted, it takes samples based on the informative function, navigates to the location, and collects samples. With each new sample, the Gaussian process regression is trained, the intensity values of the whole environment (map) are predicted, and the informativeness of each location is updated based on one of the \textbf{seven variants} in Sec.~\ref{sec:information}, which includes 6 adaptive sampling variants and one of the two baseline non-adaptive sampling variants based on the scenario.
The robot's starting location and battery timing were kept fixed for all scenarios.
The ground truth for the Wi-Fi signal map (Fig.~\ref{fig:wifi}) was generated as per the equation: $RSS = RSS_{d0} - 10n  \ln(d) + \chi_g ,$
where $RSS_{d0} = TX_{power} - 20 \space \ln  (\frac{3}{4\pi*f})$ is the signal reference power at $d0=1m$, $f$ is the signal frequency (2.4GHz), $n$ is the path loss exponent ($n=3$), $d$ is the distance between the signal source and receiver, $\chi_g$ is a Gaussian distribution with zero mean and a variance ($0.65dBm^2$) to represent fading/noise in signals, similar to the settings in \cite{adaptive}.

\subsection{Single robot experiments}
\label{sec:scenarios}

For single robot experiments, we deployed a hector UAV robot (an aerial robot) with a battery capacity to sustain 500 ROS seconds (with a real time factor close to 1) and a starting position of (4.5,0). 
To provide a thorough analysis of the impact of exploration and exploitation on online learning and mapping of the spatial distribution map, we consider two scenarios:
\begin{enumerate}
    \item Hector Trajectory (HT)
    \item Random Walk (RW)
\end{enumerate}
In Hector trajectory (HT) scenario, the UAV sweeps the whole region in a horizontal parallel strips pattern starting from bottom center of the region to upper center of the region. The idea is to make UAV familiar with the overall intensity changes of the environment. While in a random walk (RW) scenario, the UAV randomly explores the surrounding region after moving up, down, left, or right by 3 points randomly from its current position. In both scenarios, once the UAV finishes sweeping the whole region or has taken at least 15 samples from the surroundings, the informativeness of each location is updated, and the next sampling location is chosen based on one of the information function variant. The baseline variants for these two scenarios are a complete sweep trajectory of the area (for HT scenario) and a pure random walk variant (for RW scenario), respectively.

\subsection{Multi-Robot Experiments}

In multi-robot settings, researchers used Voronoi partitioning method to divide an environment for multi robot sampling \cite{kemna2017multi,AdaptiveSamplingSensorCoverage,Luo2019}. Here, the robots are driven toward the centroids of the respective Voronoi region in order to maximize the mapping (sampling) performance and minimizing the sensing cost. Robots choose the most informative location within the Voronoi region based on a utility function that encompasses both exploration and exploitation. Specifically, the work in \cite{adaptive} uses the heterogeneity of robots to weigh the Voronoi partition, which is continually updated during the sampling process. 
Motivated by the investigations in these work, to conduct an analysis in multi-robot settings, we use Voronoi partitions as a way to distribute regions among multi-robots. To divide the given region $Q\subseteq R^2$ for n robots, we divide the environment into n regions that for each robot $i$ the $V_i$ corresponds to:
\begin{equation}
    V_i = \{ q\in Q|  \lVert \mathbf{q-p_i} \rVert \leq \lVert \mathbf{q-p_j} \rVert , \forall {j \neq i} \} 
\end{equation}

In case of multi-robot sampling, we have considered the following two scenarios of Voronoi partition:
\begin{enumerate}
\item Fixed Voronoi Partition (FVP) -  Considering only the initial robot positions, the region associated with the robot is decided at the start of the experiment. We fix these positions throughout the experiment, and the utility function determines what points  within the respective region are to be chosen as target points.
\item Dynamic Voronoi Partition (DVP) - In this scenario, the Voronoi partition continuously updates as the robot moves. The target point can only be determined if it belongs to the respective partition at the time of the request based on the informative function.
\end{enumerate}

For multi-robot experiments,  $3$ simulated Jackal UGV robots were deployed to the same Gazebo simulation framework. The initial positions of the three robots are (3,2), (3,10) and (7,7), respectively. For multi-robot sampling scenario, we have only employed the random walk baseline and taken 5 random samples per robot (totaling 15 samples) within the respective Voronoi region before utilizing adaptive sampling. The baseline variant of both of these scenarios (FVP and DVP) is the random walk sampling (non-adaptive).

\subsection{Performance Metrics}
We consider the following performance metrics:
\begin{enumerate}
  \item Samples: The number of Wi-Fi signal strength samples taken by the robot using its Wi-Fi device.
  \item RMSE: The root mean squared error between the predicted mean information (Wi-Fi signal strength) through the GPR and the ground truth information. The aim is to get predictions as close as possible to the ground truth, i.e, lower RMSE.
  \item Variance: The confidence bounds of the predicted values given by the GPR. The goal is to be confident about the predicted mean value, i.e., lower variance.
  \item Cumulative Distance: Cumulative distance refers to total distance traveled by the robot. The shorter the distance traveled, the lesser power consumption.
  \item Source localization accuracy: If the location at where the maximum mean value of the predicted GP map lies within 1 m of the actual source location, then that is classified as the correct localization, else incorrect localization. The localization accuracy is then the percentage of correct localization of all trials of all source location experiments combined.
\end{enumerate}

We ran five trials per variant in each scenario. Further, the experiments were repeated for five different Wi-Fi source locations, with each being at the middle, top-left, top-right, bottom-left, and bottom-right corners of the map area. In total, we conducted 700 simulations for this analysis. 

\begin{table}
\caption{\label{tab:all} Mapping performance (mean and std) of the informative sampling functions with Hector trajectory (HT) or random walk (RW) as the baselines in single robot experiments; and fixed Voronoi partitioning (FVP) and dynamic Voronoi partitioning (DVP) in multi-robot experiments.}
\centering
\resizebox{\linewidth}{!}{%
\begin{tabular}{|c|c|c|c|c|}
\hline
 \multicolumn{5}{c}{Single robot experiments} \\
\hline
HT Scenario & Samples      & RMSE         & Variance       & Cumulative Distance \\ \hline
Alpha75         & 219 ±  4     & 4.27 ±  0.15 & 11.23 ±  9.93  & 93.15 ±  25.59      \\ \hline
Alpha50         & 207 ±  8.66  & 4.19 ±  0.09 & 13.45 ±  10.97 & 168.33 ±  60.57     \\ \hline
Alpha25         & 182 ±  7.42  & 4.13 ±  0.08 & 5.56 ±  0.51   & 365.95 ±  63.34     \\ \hline
MaxVar          & 153 ±  2     & 4.18 ±  0.17 & 2.48 ±  0.33   & 618.33 ±  16.78     \\ \hline
MaxMean         & 222 ±  0     & 4.71 ±  0.49 & 32.09 ±  46.63 & 78.63 ±  4.9        \\ \hline
MaxVarMaxMean   & 177 ±  15.39 & 4.24 ±  0.17 & 4.92 ±  0.97   & 425.45 ±  115.94    \\ \hline
RW              & 36 ±  1      & 4.82 ±  0.66 & 7.12 ±  2.13   & 68.32 ±  0.64       \\ \hline
\hline
\hline
RW Scenario & Samples      & RMSE         & Variance       & Cumulative Distance \\ \hline
Alpha75         & 205 ±  14.66 & 4.33 ±  0.35 & 26.27 ±  14.27 & 175.2 ±  75.87      \\ \hline
Alpha50         & 194 ±  12.61 & 4.23 ±  0.1  & 16.13 ±  10.27 & 284.8 ±  91.59      \\ \hline
Alpha25         & 168 ±  7.68  & 4.11 ±  0.07 & 5.73 ±  0.57   & 494.11 ±  70.57     \\ \hline
MaxVar          & 145 ±  3.46  & 4.18 ±  0.1  & 2.49 ±  0.29   & 696.28 ±  16.86     \\ \hline
MaxMean         & 226 ±  1     & 6.06 ±  1.1  & 37.84 ±  44.57 & 61.44 ±  4.94       \\ \hline
MaxVarMaxMean   & 170 ±  10.63 & 4.2 ±  0.11  & 4.97 ±  0.54   & 489.65 ±  89.8      \\ \hline
RW              & 150 ±  1     & 4.45 ±  0.22 & 3.1 ±  0.94    & 520.94 ±  6.34      \\ \hline
\hline
 \multicolumn{5}{c}{Multi-robot experiments} \\
\hline
FVP Scenario & Samples      & RMSE         & Variance      & Cumulative Distance \\ \hline
Alpha75        & 426 ±  34.81 & 4.11 ±  0.12 & 9.78 ±  2.82  & 149.12 ±  39.72     \\ \hline
Alpha50        & 403 ±  35.76 & 4.05 ±  0.1  & 6.52 ±  1.69  & 224.74 ±  58.42     \\ \hline
Alpha25        & 361 ±  31.51 & 4.02 ±  0.1  & 3.87 ±  0.68  & 332.02 ±  60.44     \\ \hline
MaxVar         & 236 ±  5     & 4.02 ±  0.07 & 1.87 ±  0.13  & 592.91 ±  8.69      \\ \hline
MaxMean        & 445 ±  44.78 & 4.97 ±  0.91 & 17.07 ±  5.91 & 68.8 ±  16.94       \\ \hline
MaxVarMaxMean  & 377 ±  29.65 & 4.06 ±  0.08 & 4.2 ±  0.23   & 250.19 ±  28.6      \\ \hline
RW             & 204 ±  6.24  & 4.77 ±  0.46 & 4.74 ±  1.3   & 514.62 ±  5.96      \\ \hline
\hline
\hline
DVP Scenario & Samples      & RMSE         & Variance       & Cumulative Distance \\ \hline
Alpha75        & 463 ±  51.11 & 4.16 ±  0.15 & 10.1 ±  1.98   & 89.6 ± 16.35        \\ \hline
Alpha50        & 408 ±  39.29 & 4.1 ±  0.1   & 8.06 ±  0.97   & 120.42 ± 11.84      \\ \hline
Alpha25        & 382 ±  34.22 & 4.08 ±  0.1  & 4.5 ±  0.47    & 219.16 ± 29.98      \\ \hline
MaxVar         & 244 ±  10.72 & 4.01 ±  0.07 & 1.86 ±  0.27   & 538.42 ± 47.83      \\ \hline
MaxMean        & 444 ±  29.83 & 4.55 ±  0.55 & 24.08 ±  34.05 & 62.94 ± 11          \\ \hline
MaxVarMaxMean  & 408 ±  37.96 & 4.07 ±  0.09 & 4.34 ±  0.27   & 233.89 ± 29.84      \\ \hline
RW             & 211 ±  20.95 & 4.65 ±  0.39 & 3.71 ±  1.01   & 485.69 ± 72.39      \\ \hline
\end{tabular}
}
\end{table}

\begin{table}
\caption{\label{tab:localization} Source localization accuracy (\%) by all approaches.}
\centering
\resizebox{\linewidth}{!}{%
\begin{tabular}{|c|c|c|c|c|c|c|c|c|c|}
\hline
                                             &                                   & \textbf{Samples}   & \textbf{Alpha75} & \textbf{Alpha50} & \textbf{Alpha25} & \textbf{MaxVar} & \textbf{MaxMean} & \textbf{MaxVarMaxMean} & \textbf{RW/HT} \\ 
                                             \hline
\multirow{14}{*}{\textbf{\begin{turn}{90}Multi-robot cases\end{turn}}} 

                                             & \multirow{7}{*}{\textbf{HT}} & 10                 & 24               & 32               & 28               & 28              & 32               & 28                     & 24             \\   
                                             &                                   & 25                 & 40               & 32               & 24               & 40              & 32               & 32                     & 28             \\   
                                             &                                   & 35                 & 84               & 68               & 68               & 72              & 68               & 76                     &     -           \\   
                                             
                                              &                                   & 45          &100	&100	&100	&92	&72	&96  &    -      \\

                                             &                                   & 50                 & 100              & 100              & 100              & 92              & 72               & 92                     &     -           \\   
                                             &                                   & After half samples               & 100              & 100              & 100              & 96              & 72               & 96                     & 48             \\   
                                             &                                   & After last sample                & 100              & 100              & 100              & 100             & 72               & 100                    & 68             \\ \cline{2-10}

                                             & \multirow{6}{*}{\textbf{RW}} & 10                 & 56               & 48               & 44               & 60              & 52               & 56                     & 40             \\   
                                             &                                   & 25                 & 96               & 96               & 96               & 100             & 56               & 100                    & 64             \\   
                                             &                                   & 35                 & 100              & 96               & 96               & 100             & 56               & 100                    & 80             \\

                                              &                                   & 45
                                              &100	&96	&100	&100	&56	&100	&80

                                                     \\  
                                             
                                             &                                   & 50                 & 100              & 100              & 100              & 100             & 56               & 100                    & 80             \\   
                                             &                                   & After half samples & 100              & 100              & 100              & 100             & 56               & 100                    & 80             \\   
                                             &                                   & After last sample  & 100              & 100              & 100              & 100             & 56               & 100                    & 100            \\  \hline      
                                       
\multirow{14}{*}{\textbf{\begin{turn}{90}Multi-robot cases\end{turn}}} 

& \multirow{7}{*}{\textbf{FVP}}     & 10                 & 40               & 40               & 32               & 36              & 36               & 48                     & 48             \\   
                                             &                                   & 25                 & 96               & 100              & 96               & 96              & 44               & 100                    & 64             \\   
                                             &                                   & 35                 & 100              & 100              & 100              & 100             & 56               & 100                    & 64             \\   
                                              &                                   & 45    
                                              
                                              &100	&100	&100	&100	&64	&96	&60
                                              
                                                      \\

                                             &                                   & 50                 & 100              & 100              & 100              & 100             & 64               & 96                     & 60             \\   
                                             &                                   & After half sample  & 100              & 100              & 100              & 100             & 76               & 100                    & 60             \\   
                                             &                                   & After last sample  & 100              & 100              & 100              & 100             & 76               & 100                    & 60             \\ \cline{2-10} 
                                             & \multirow{7}{*}{\textbf{DVP}}     & 10                 & 36               & 40               & 28               & 32              & 32               & 40                     & 48             \\   
                                             &                                   & 25                 & 100              & 100              & 100              & 92              & 56               & 88                     & 56             \\   
                                             &                                   & 35                 & 100              & 100              & 100              & 96              & 76               & 100                    & 56             \\   
                                             
                                             &                                   & 45         
                                             &100	&100	&100	&96 &88	&100	&60

                                              \\   
                                             
                                             &                                   & 50                 & 100              & 100              & 100              & 96              & 92               & 100                    & 60             \\   
                                             &                                   & After half samples  & 100              & 100              & 100              & 100             & 96               & 100                    & 60             \\   
                                             &                                   & After last sample  & 100              & 100              & 100              & 100             & 96               & 100                    & 68             \\ \hline
\end{tabular}
}
\end{table}

\section{Results and Analysis}
\label{sec:results}

\begin{figure}[ht]
\centering
\subfigure[Single robot - Hector Trajactory (HT) Scenario]{
\includegraphics[width=0.32\columnwidth]{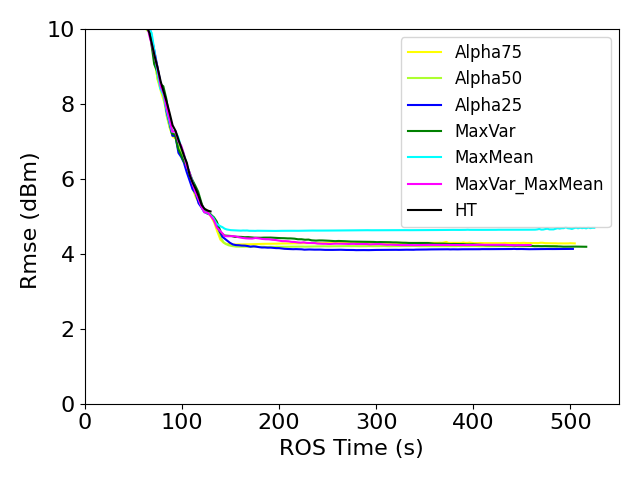}
\includegraphics[width=0.32\columnwidth]{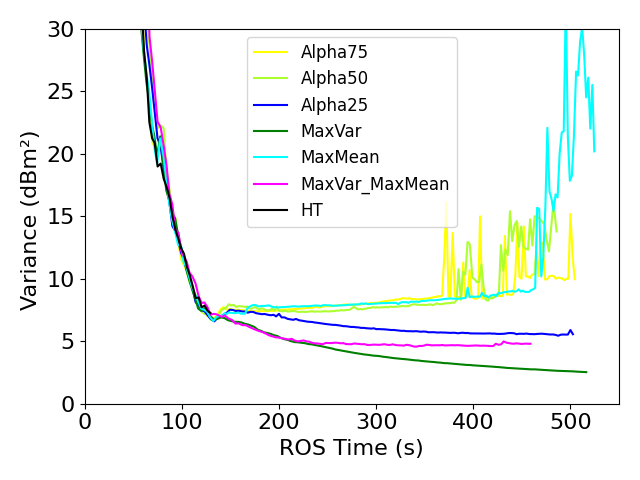}       
\includegraphics[width=0.32\columnwidth]{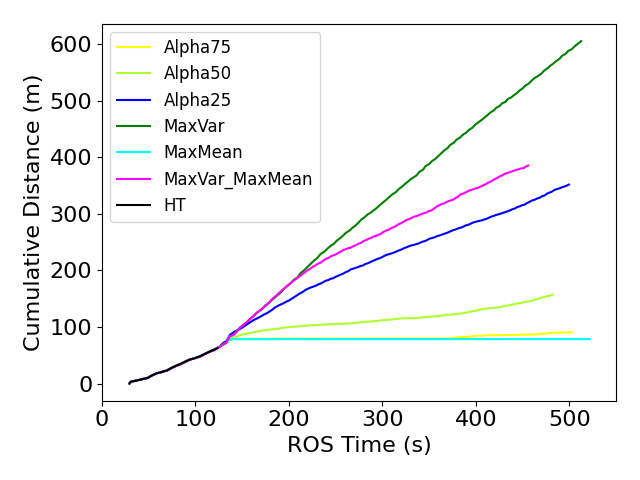}    
\label{fig:HT}
}

\subfigure[Single robot - Random Walk (RW) trajactory Scenario]{
\includegraphics[width=0.32\columnwidth]{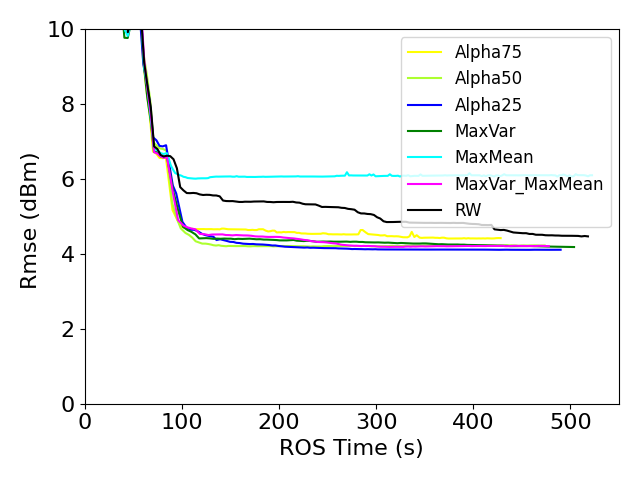}
\includegraphics[width=0.32\columnwidth]{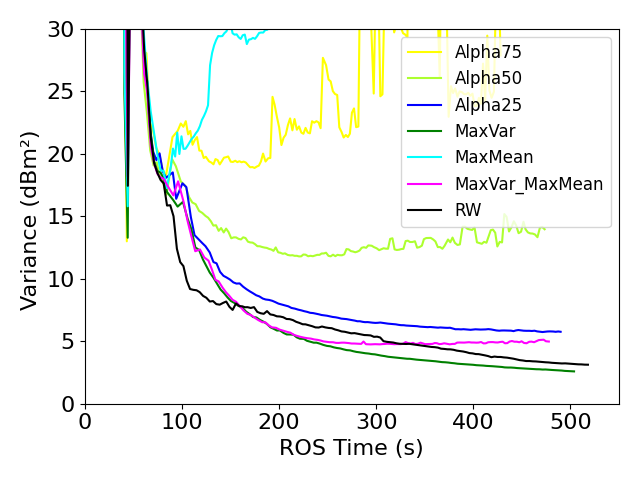}       
\includegraphics[width=0.32\columnwidth]{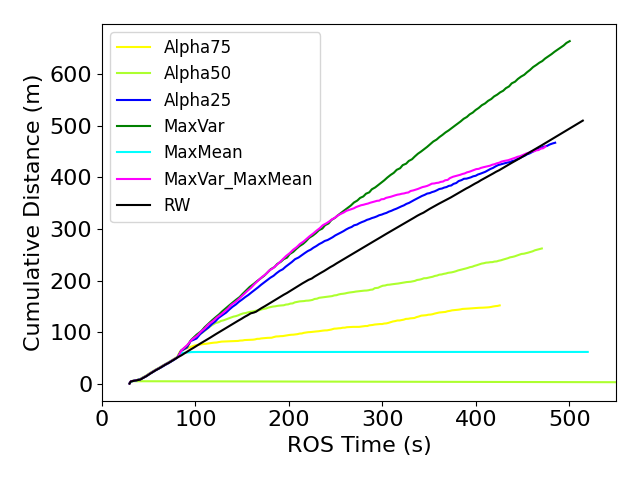}    
\label{fig:RW}
}

\subfigure[Multi-robot - Fixed Voronoi Partitioning (FVP) Scenario]{
\includegraphics[width=0.32\columnwidth]{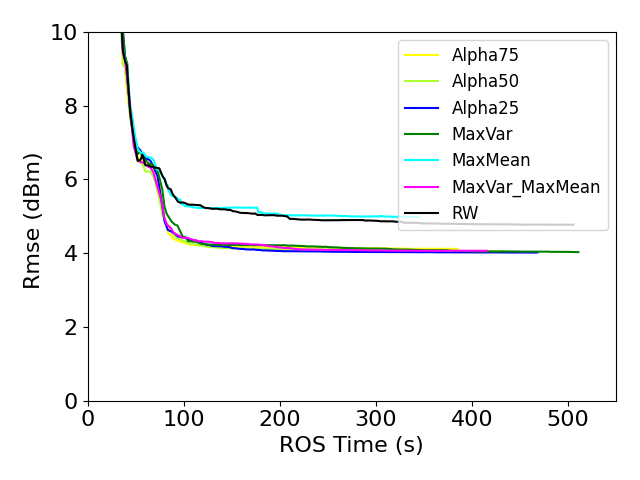}
\includegraphics[width=0.32\columnwidth]{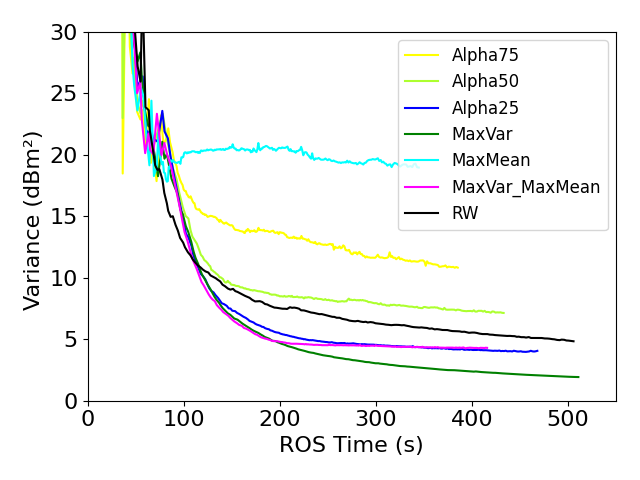}       
\includegraphics[width=0.32\columnwidth]{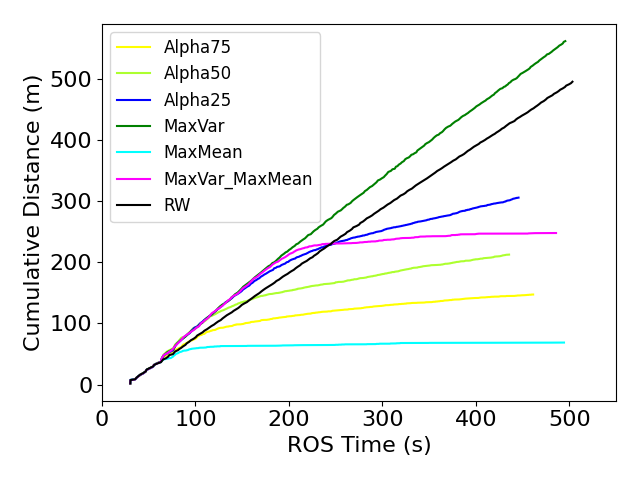}    
\label{fig:FVP}
}

\subfigure[Multi-robot - Dynamic Voronoi Partitioning (DVP) Scenario]{
\includegraphics[width=0.32\columnwidth]{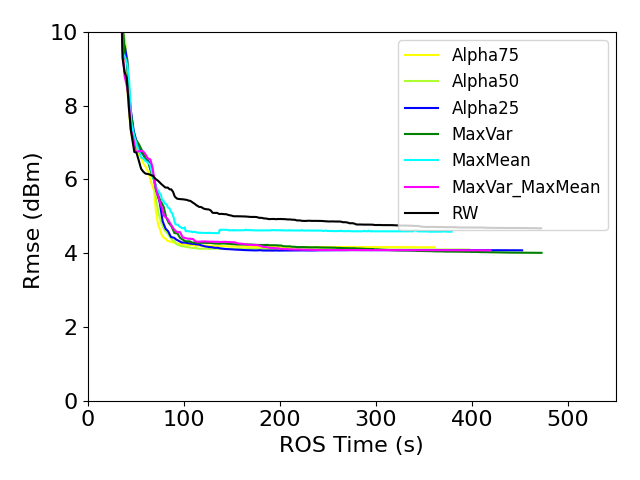}
\includegraphics[width=0.32\columnwidth]{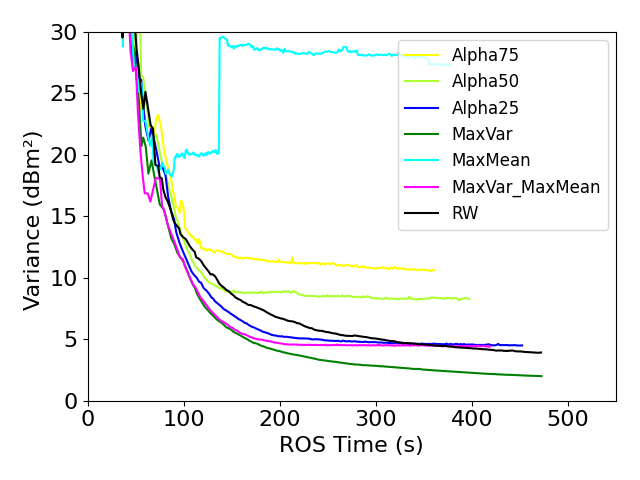}       
\includegraphics[width=0.32\columnwidth]{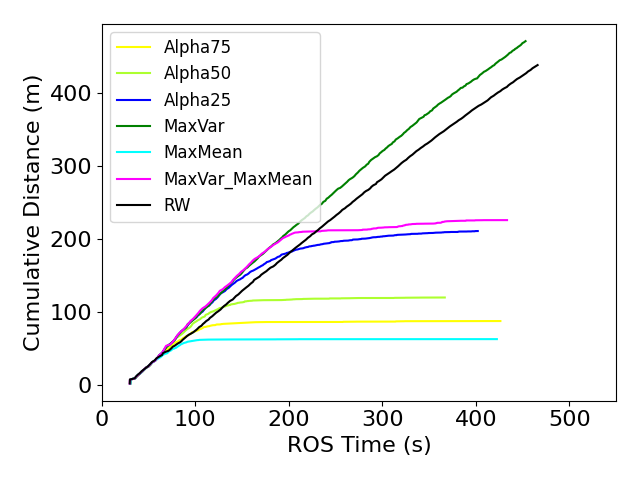}    
\label{fig:DVP}
}

\caption{RMSE (left), Variance (center) and Cumulative Distance (right) for different information functions in single robot and multi-robot experiments.}
\label{fig:performance}
\end{figure}

Table \ref{tab:all} provides a summary of the performance metrics results obtained by averaging the data collected over all trials with different source locations.  
Fig.~\ref{fig:performance} shows the plots for the three key performance metrics: RMSE, Variance, and cumulative distance with respect to time in all scenarios.

In general, exploration seeks to minimize variance and RMSE of the predicted map, while exploitation seeks to make use of the predictions as soon as possible (e.g., leading to identify the source location, i.e., the place with the maximum signal intensity). An informative function can be either exploitation-based, exploration-based, or a weighted combination of both. In our variants, the MaxMean strategy does pure exploitation, whereas MaxVar, Hector trajectory, and random walk are pure exploration strategies. The rest of the variants combine exploration and exploitation.

\subsection{Exploitation Perspective}
We take the source localization example as the objective of exploitation in our work. To properly locate a source, a robot should detect provide a GP map with maximum mean at a point within one meter away from the real source location in any direction.
Table \ref{tab:localization} shows the source localization accuracy of all variances in all scenarios based on the number of times the resultant GP map can be used to correctly identify the source locations after collecting different numbers of samples during the experiments. 

\paragraph{Single robot cases}
It can be observed that the localization accuracy of each approach improved with increasing numbers of samples. 
We are interested in identifying approaches that allow us to obtain better results with fewer observations. The RW, MaxMean and HT approaches did not perform well, especially in the early stages of the experiment. With HT, fewer measurements are taken, while with MaxMean, the measurements are taken repetitively at the same location (local maxima) since the information function (with $\alpha=1$) is only dependent on the predicted GP mean. 

In general, pure RW approach had improved performance, but it was slower than the other approaches.  RW variants' results were better after just 25 samples than HT's after 35 samples. After 45 samples, the performance was almost similar.
We found that the Alpha50 approach to be unsuitable for exploitation. In RW scanerios where the initial 15 samples were taken based on random walk, the best variants were MaxVar, Alpha75, MaxVarMaxMean, and Alpha25. Alpha25 and MaxVar, however, do not represent cost-effective approaches since both involve very long distances. As a result, MaxVarMaxMean and Alpha75 both converge faster and are cost-effective. If MaxVarMaxMean approach doesn't satisfy the variance and RMSE threshold, it is same as MaxVar. We see that the Alpha75 as best approach when minimizing distance cost is the first priority (e.g., if energy availability is heavily limited \cite{parasuraman2012energy}). Alpha25 works best in situations where the source localization has to be even more accurate and cost-effectiveness is less of a concern. When cost is not a concern at all, then surprisingly, the MaxVar approach ($\alpha=0$) provided the best exploitation performance in satisfying both mapping accuracy and confidence.

\paragraph{Multi-robot cases}
We observe similar results for the multi-robot cases compared to the single robot cases, where Alpha25 variant best balances both exploitation objective and the cost requirements. The improvements in performance found with MaxMean and RW were slower than single robot cases and did not reach the same level as other approaches. After 25 samples, the dynamic and fixed Voronoi partitions performed close to each other, and they were successful at locating the source much faster than the single robot experiments even with just 25 samples.

\subsection{Exploration Perspective}
For exploration, the objective is to obtain accurate predictions of the sampled environmental process with the highest confidence bounds in all areas of the map.

\paragraph{Single-robot results}
Here, the MaxMean resulted in the highest RMSE, followed by all the other methods, but with the worst confidence (variance). As variants of both baselines, MaxVar, MaxVarMaxMean, and Alpha25 performed exceptionally well, with MaxVar achieved the best convergence on the variance and RMSE together. The RW baseline also performed well, but it is worth noting that the RW method was the last one to converge. MaxMean and Alpha75 approaches always chose the same target position, giving them identical samples, which reduced variance. As HT baseline yields only a few samples, it performed poorly in terms of variance.

We can safely deduce that MaxVar (with $\beta=1$) is the best approach when it comes to exploration. However, the cumulative distance is also the greatest for this approach. Furthermore, MaxVar works better with the random walk scenario. Alpha25 provided the best exploration performance when cost and efficiency are also considered. We also found that higher values of the coefficient $\beta$ would give better exploration results, as expected. 

\paragraph{Multi-robot cases}
Distance plots for multi-robots scenarios show that all variants of DVP approaches took much shorter travel distances than the same approaches based on FVP. However, the FVP scenario resulted in improved variance as well as speed of convergence compared to DVP scenarios. 
Consequently, we can conclude that DVP scenario is suitable for cost-effective sampling (less energy), while the FVP scenario is suited for faster convergence and better exploration results. While MaxVar has promising results in terms of variance, and has taken fewer samples than Alpha75 and MaxVarMaxMean, its distance cost is almost twice that of Alpha75 and MaxVarMaxMean. Accordingly, when comparing convergence, number of samples, and travel costs together, FVP in combination with Alpha75 or MaxVarMaxMean variants outperformed other variants. 

\subsection{Further Discussions}
When analyzing the results from the perspectives of exploration vs. exploitation sampling objectives, MaxVar variant produced best results by making the robot traveling throughout the map. However, the Alpha25 stood out as an optimum variant that balances the performance in terms of RMSE, Variance, efficiency, and source localization together, making it suitable to both exploration and exploitation objectives. 

More detailed data of the results and high-resolution figures are available as an appendix to this paper at \url{http://hero.uga.edu/research/adaptivesampling/}.

\paragraph{Impact of source locations on the sampling performance}
We also analyzed the impact of different source locations on the sampling performance (results for these special cases are available in Appendix). We found that there was almost no impact on the results across all sources, especially  when the $\beta$ value (i.e., the weight towards confidence bounds) is higher. However, for variants where $\alpha$ the value is higher (MaxMean, Alpha0.75, and Alpha0.5), they gave significantly different results for the furthest source locations bottom-right (0,14) and top-right (9,14) of the map area. This could be attributed to that fact that when $alpha$ is higher, exploitation is more preferred and therefore localizing a source that is much farther could be difficult to accomplish. 
In summary, the effect of source locations was not observed for informative functions with greater weights for variance (exploration).

\paragraph{Impact of Wi-Fi signal distribution on the sampling performance}
Further, we analyzed the impact of the signal distribution itself on the results by repeating all the single robot cases with different path loss exponent (n=2) (results for these special cases are included in the Appendix). 
It was observed that all approaches with n=2 performed quite well in comparison to the cases of n=3, and we found that the change in both variance and RMSE metrics were smoother for all variants when n=2, than the same approaches for n=3. Nevertheless, the change in the signal distribution had minimal impact on our analysis, and the observations made for n=3 above hold for n=2 as well.

\clearpage 
\section{Conclusion}
This study provided an understanding of how exploration and exploitation objectives differ in mapping accuracy and efficiency changed for various variants of informative functions in adaptive sampling of environmental sensing or physical processes. For two baseline approaches, we used the Gaussian processes based mapping and evaluated variants of utility function on simulated Wi-Fi signal map by extensively analyzing on single robot (UAV) based experiments and multi-robot (UGV) based experiments. Our results show that approaches with higher weights to confidence bounds provide the best performance results when seeking the next sampling points. The analysis from our data and the results provide insights in designing optimal information functions for achieving adaptive information sampling.

\bibliographystyle{IEEEtran}
\bibliography{ref}

\end{document}